\documentclass[10pt,twocolumn,letterpaper]{article}

\usepackage{wacv}
\usepackage{times}
\usepackage{epsfig}
\usepackage{graphicx}
\usepackage{amsmath}
\usepackage{amssymb}
\usepackage{booktabs}
\usepackage{appendix}

\def\wacvPaperID{8} 

\wacvapplicationstrack 

\wacvfinalcopy 

\ifwacvfinal
\usepackage[breaklinks=true,bookmarks=false]{hyperref}
\else
\usepackage[pagebackref=true,breaklinks=true,colorlinks,bookmarks=false]{hyperref}
\fi

\begin{document}

\title{\LARGE \bf
\textit{SeaDroneSim}: Simulation of Aerial Images for Detection of Objects Above Water}

\author{Xiaomin Lin$^{1}$, Cheng Liu$^{1}$, Allen Pattillo$^{2}$, Miao Yu$^{3}$, Yiannis Aloimonous$^{1}$\\
$^{1}$Perception and Robotics Group, University of Maryland - College Park, MD\\
$^{2}$ University of Maryland Extension, MD\\
$^{3}$ Mechanical Engineering, University of Maryland - College Park, MD\\
{Email:\tt\small \{xlin01, cliu2021, dapatt, mmyu, jyaloimo\}@umd.edu  }
}

\maketitle
\thispagestyle{empty}

\begin{abstract}

Unmanned Aerial Vehicles (UAVs) are known for their speed and versatility in collecting aerial images and remote sensing for land use surveys and precision agriculture. With UAVs' growth in availability and accessibility, they are now of vital importance as technological support in marine-based applications such as vessel monitoring and search-and-rescue (SAR) operations. High-resolution cameras and Graphic processing units (GPUs) can be equipped on the UAVs to effectively and efficiently aid in locating objects of interest, lending themselves to emergency rescue operations or, in our case, precision aquaculture applications. Modern computer vision algorithms allow us to detect objects of interest in a dynamic environment; however, these algorithms are dependent on large training datasets collected from UAVs, which are currently time-consuming and labor-intensive to collect for maritime environments. 

To this end, we present a new benchmark suite, \textit{\textbf{SeaDroneSim}}, that can be used to create photo-realistic aerial image datasets with ground truth for segmentation masks of any given object. Utilizing only the synthetic data generated from \textit{\textbf{SeaDroneSim}}, we obtained 71 a mean Average Precision (mAP) on real aerial images for detecting our object of interest, a popular, open source, remotely operated underwater vehicle (BlueROV) in this feasibility study. The results of this new simulation suit serve as a baseline for the detection of the BlueROV, which can be used in underwater surveys of oyster reefs and other marine applications.

Index Terms—\textbf{simulation, aerial images, automation, BlueROV,  detection}. 
\vspace{-5mm}
\end{abstract}
\begin{figure}
 \centering
{\includegraphics[width=0.5\textwidth]{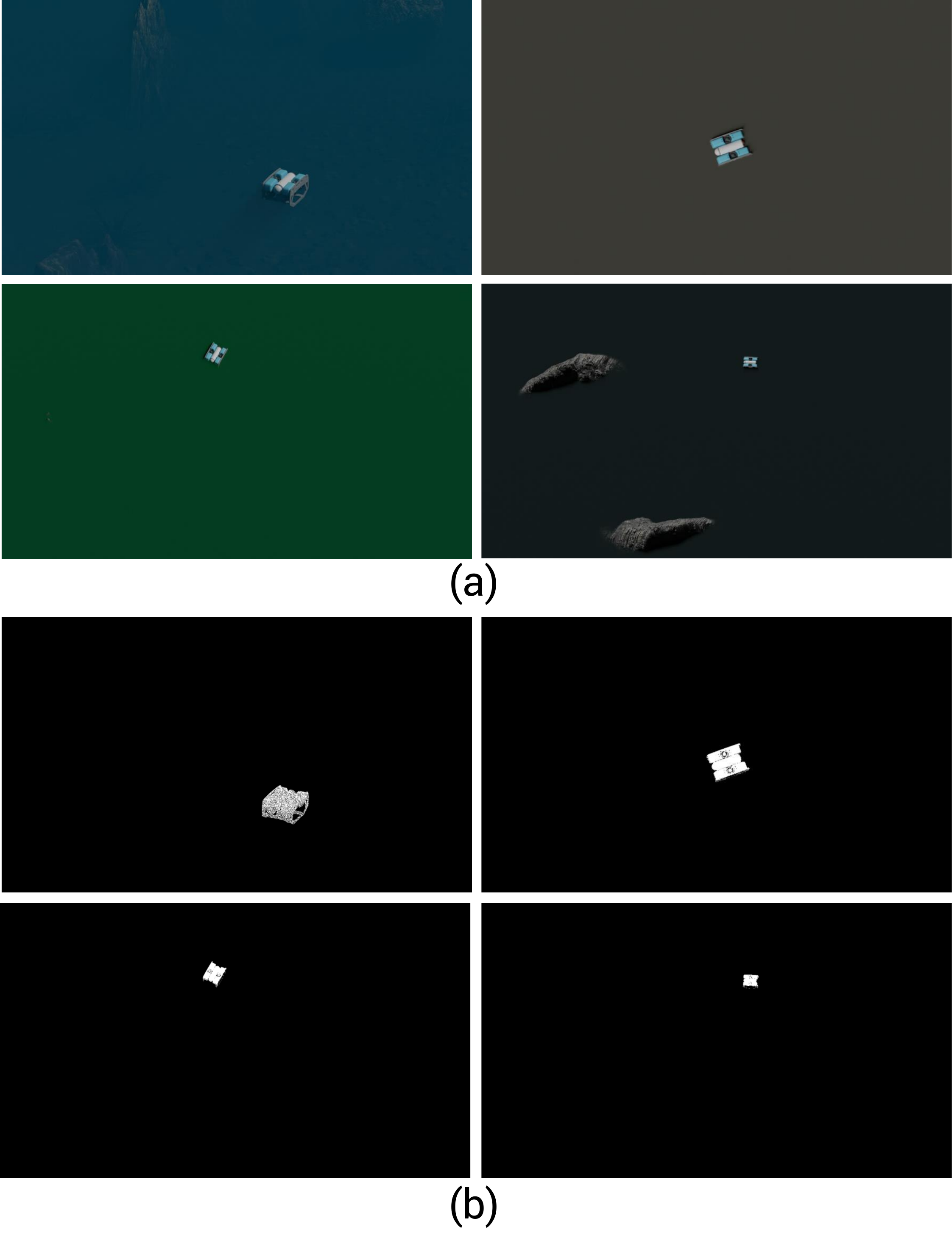}}
\caption{(a) Typical image examples with varying altitudes, angles of view and background water color. (b) Corresponding ground truth masks for the object of interest (BlueROV).}
\vspace{-5mm}
\label{fig:sample_images}
\end{figure}
\section{Introduction}
\begin{figure*}[ht!]
\includegraphics[width=\textwidth]{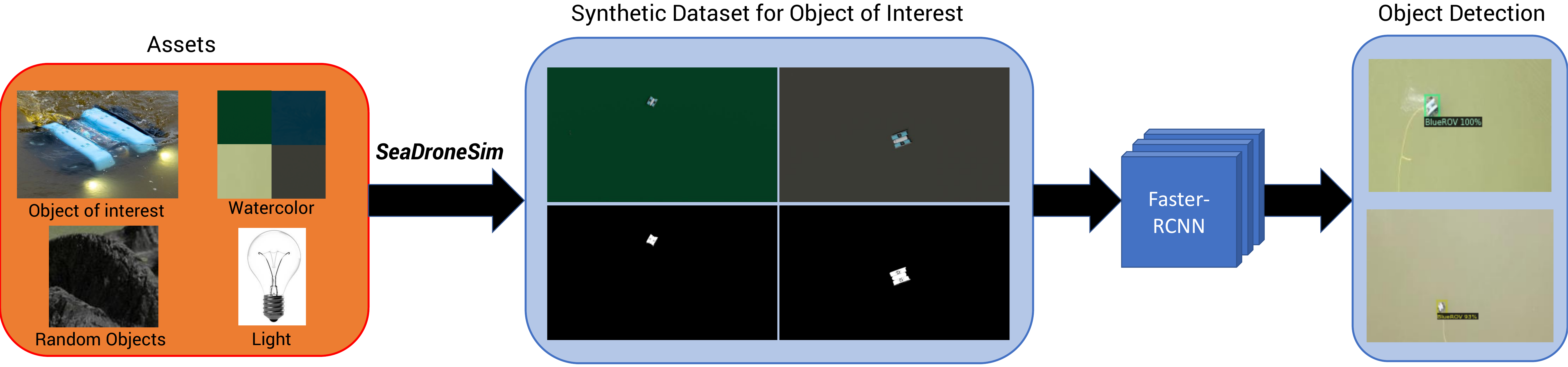}
\centering
\vspace{-3mm}
\caption{An overview of our approach: 3D models for the object of interest are used in \textit{SeaDroneSim} to generate Synthetic datasets. Noted the synthetic dataset would include its ground truth mask for the object of interest. The synthetic dataset generated is then fed into a Neural network to obtain the object detection result. Note: Object Detection images are cropped and enlarged for better visualization.}
\label{fig:overview}
\vspace{-3mm}
\end{figure*}
Unmanned Aerial Vehicles (UAVs) equipped with high-resolution cameras and Graphics Processing Units (GPUs) are increasingly used in a variety of tasks, ranging from land-use classification \cite{chen2020imaging, akar2017mapping}, precision agriculture \cite{raj2020precision,tsouros2019review, chen2020imaging}, coastal management \cite{adade2021unmanned}, crowd surveillance \cite{de2019multi,motlagh2017uav}, and recently in assisting with search and rescue (SAR) missions \cite{albanese2021sardo,cacace2016multimodal,gallego2019detection,karaca2018potential, mishra2020drone} and post-disaster exploration missions by providing a bird's eye view over the scene. Specifically, in marine scenarios, where a large ocean area needs to be searched quickly, UAVs provide vital aid to emergency search and rescue operations \cite{yeong2015review}. For SAR missions, a robust vision-based system is needed to locate/track objects of interest in maritime environments with different lighting conditions, image altitudes, viewing angles, and water colors. Currently, most of these are vision-based, data-driven systems employing deep neural networks. These systems are dependent on large datasets to correctly recognize objects under diverse environmental conditions. However, there are only a few datasets available, which are limited in their size and variety of objects of interest. The process of collecting large-scale datasets for specific object detection tasks is slow, time-consuming, and has poor scalability. Using a manual approach, only a few objects of interest can only be located/tracked within a specific dataset for that task. Varga \cite{varga2022seadronessee} proposed a relatively large dataset for the Maritime environment, yet with a nearly infinite number of potential objects of interest there is a need for methods to rapidly generate large-scale datasets in maritime environments for a variety of objects, including our object of interest, the remotely operated underwater vehicle - BlueROV Fig.~\ref{fig:ROV}.

Similar to Kiefer et al. \cite{kiefer2021leveraging}, this work aims to help fill the gap of large-scale maritime-based datasets captured from UAVs. We introduce a simulation suit for generating simulated images of maritime-based aerial images, called \textit{\textbf{SeaDronesSim}}. We are capable of generating videos and images of any object of interest with a 3D model Fig.~\ref{fig:ROV} in open water with various lighting conditions, image altitudes, viewing angles, and background water colors (see Fig.~\ref{fig:sample_images}(a)). To fully utilize the \textit{SeaDronesSim}, the ground truth for the object of interest is generated at the same time (see Fig.~\ref{fig:sample_images}(b)). Most SAR missions require a vision-based system to be able to detect and track from a great distance. \textit{SeaDronesSim} is able to vary its resolution for the Red, Green, and Blue (RGB) footage captured. To study the feasibility of object recognition purely from synthetic images generated from \textit{SeaDronesSim}, we used our object of interest, the BlueROV, for which we have a detailed 3D computer model on hand for simulating various image poses from Patrick\cite{patrickelectric2018}. Moreover, the simulation provides all the ground truth information of the aerial camera (altitude taken, rotation) and the BlueROV (rotation), allowing us to output all the metadata for the aerial camera and the objects of interest. We believe all this metadata could be used in the future to develop multi-modal systems to improve accuracy and speed.
Finally, we conducted extensive experiments to compare the datasets that we generated using a state-of-the-art model and established baseline models for BlueROV detection. These results serve as a starting point for our \textit{SeaDronesSim} simulation suit. 
To streamline the process of object detection, our goal is to utilize  advancements in robotics and artificial intelligence that enable us to gather UAV-based aerial images from the maritime environment to automate the process of object detection and tracking.
\textit{SeaDronesSim} is among the first simulations for UAV-based aerial images from the maritime environment \cite{kiefer2021leveraging}. Our main contributions are as follows:
\begin{itemize}
\item We propose a novel simulation suit for generating UAV-based aerial images for the maritime environment.
\item To the best of our knowledge, we established a baseline for detecting BlueROV in open water.  
\item We open-source \footnote{\url{https://github.com/prgumd/SeaDroneSim}}our \textit{SeaDronesSim} and dataset associated with this work to accelerate further research.
\item We proposed a pipeline for autonomously generating UAV-based maritime images for objects of interest and detecting them. 
\end{itemize}

The rest of this paper is organized as follows: We first place this work in the context of related works in Sec. ~\ref{section:related_word}. Then, we describe the proposed simulation which is used to create realistic images in Sec. ~\ref{section:proposed_approach}. We then present some quantitative and qualitative evaluations of our approach in Sec. ~\ref{section:experiments_and_results}. We conclude our work in Sec. ~\ref{section:conclusion} with parting thoughts on future work.

    

\section{Related work}
\label{section:related_word}
In this Section, we first review some of the major datasets for aerial images and maritime environments. Then we describe some of the simulations for the maritime/aerial domain.  
\subsection{Datasets for Maritime Environments}
Large-scale datasets are necessary when developing modern computer vision algorithms. 
However, many datasets for maritime environments are satellite-based synthetic aperture radar imagery and are used for remote sensing tasks. In 2018, Airbus \cite{kaggle_2018} released 40k satellite images (with instance segmentation labels) from synthetic aperture radars for ship detection. Gallego\cite{gallego2018automatic} also proposed a method for autonomous ship detection with aerial images.
Li et al. \cite{li2018hsf} proposed another dataset for ships consisting mainly of Google Earth and some UAV-based images. In 2019, Lygouras et al. \cite{lygouras2019unsupervised} worked on human detection with UAV-based images. However, Lygouras' dataset was limited, as the dataset is collected either near-shore or in a swimming pool. Keifer et al. \cite{kiefer2021leveraging} analyzed real and synthetic maritime and terrestrial images from both satellite and UAV sources for detecting boats and people. Another UAV-based image dataset, released by Varga \cite{varga2022seadronessee}, is tailored toward recognizing objects in the water and collecting large-scale data for the maritime environment. However, Varga's dataset only provides six classes of objects. There are many more objects of interest in the SAR tasks such as hovercraft, floating planes, humans waving for help, etc. There is no dataset that can represent all objects of interest for every conceivable detection task, thus the need for an automated process to generate this data synthetically as needed. 
\subsection{Simulation}
Instead of collecting large-scale datasets for detection tasks, we follow the conceptual approach of detecting oysters \cite{lin2022oystersim}\cite{lin2022oysternet} and propellers \cite{sanket2021prgflow} which utilizes a 3D model of the object to generate an enormous amount of data synthetically. In particular, we used a 3D model of the underwater BlueROV robot to create a maritime dataset for BlueROV detection.

When there is a lack of large-scale datasets for some robotics tasks, different research groups have developed various simulations to meet their needs. Most simulators for the aerial and maritime domains focus on controlling the drone for safety operations \cite{cox2007use} and rapid control \cite{velasco2020open}. Abujob \cite{abujoub2018unmanned} used simulation to verify his algorithm of landing the drone on the ship with the prediction of ship motion.
The most similar simulation to \textit{SeaDroneSim} is from Matlab UAV Toolbox \cite{matlabuav} which provides control of the drone and also images from a camera attached to the drone. However, this work focuses on controlling the drone for terrestrial applications and does not provide segmentation ground truth for the object of interest. 

With the ultimate goal of developing an autonomous UAV-based surveillance system, we acknowledge the need for object detection methods. 
Moreover due to the lack of large-scale datasets for the maritime environment and the lack of literature addressing this problem, we are seeking an alternative path for generating maritime datasets through synthetic image generation.
To the best of our knowledge, we are among the first to propose a simulator (\textit{SeaDroneSim}) for maritime environment datasets and use that data to recognize objects of interest.
\begin{figure}[t!]
\includegraphics[width=\linewidth]{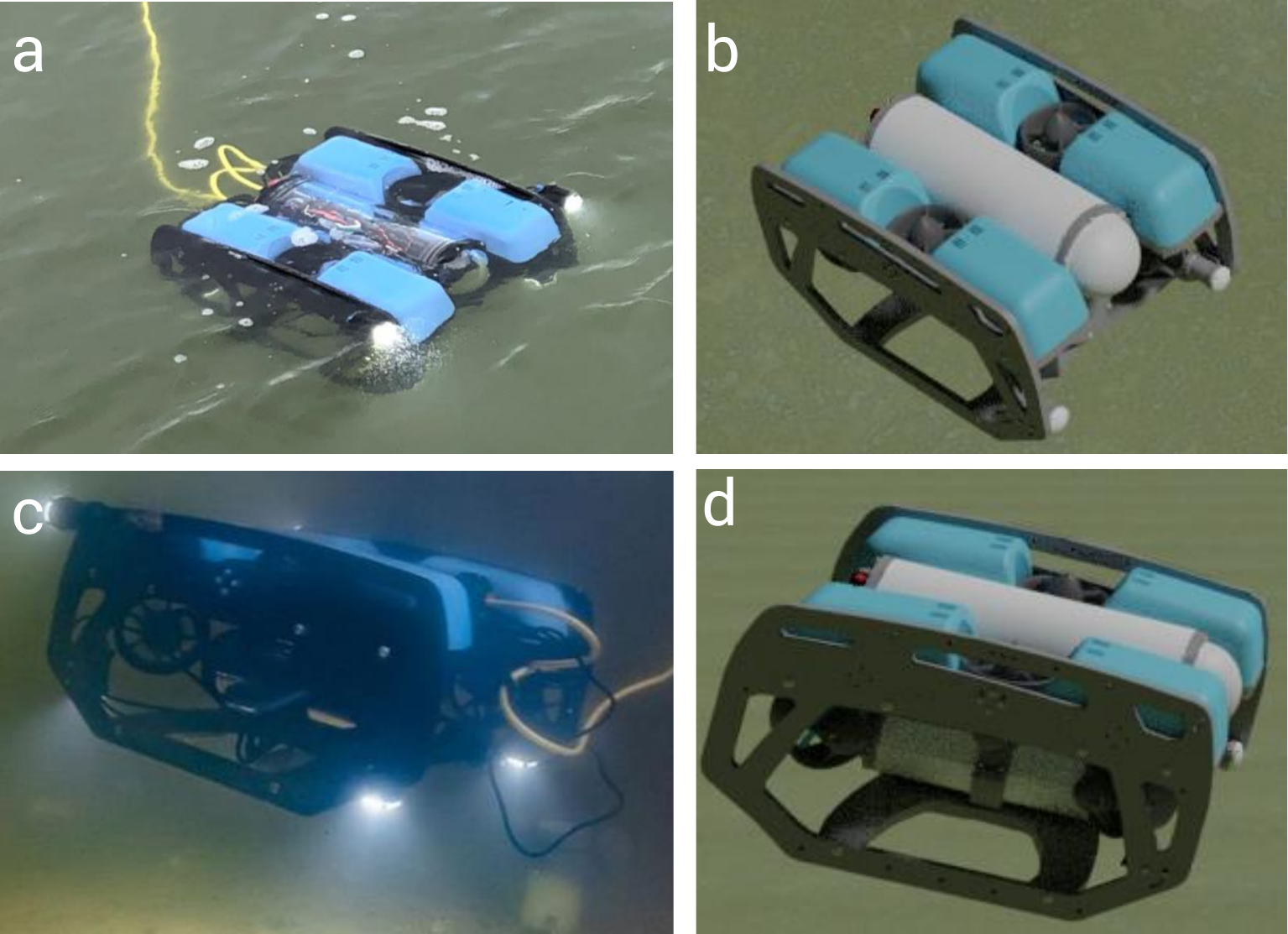}
\centering
\caption{(a) BlueROV in water near Horn Point Lab, (b) BlueROV in Blender$^{\text{TM}}$, (c) BlueROV in water in Horn Point Lab, (d) BlueROV in Blender$^{\text{TM}}$.}
\label{fig:ROV}
\vspace{-3mm}
\end{figure}
\textit{SeaDroneSim} will be described in the next Section.

\section{Proposed Approach}
\label{section:proposed_approach}
In this work, we propose a novel simulation built from Blender$^{\text{TM}}$ \cite{blender} game engine. Then we used the simulation for creating UAV-based maritime images which are then used to train a Neural Network for object detection. As we can see in Fig.~\ref{fig:overview}), we first choose our object of interest and different assets(i.e. water color, light, random objects) and input them into the simulation engine that we build. Moreover, utilizing the data produced from the simulation engine, we can come up with a detection network for detecting objects of interest.    In this section, we will go through some of the details of the implementation of the \textit{SeaDronesSim}.
\subsection{Object of Interest}
It is critically important to have an accurate 3D model of the object for this work. Only with an accurate 3D model of the object in the simulation, can we correctly represent the visual appearance from the UAV-based maritime images. In this work, we select BlueROV as our object of interest because of the availability of its 3D model\cite{patrickelectric2018} and actual object for collecting real datasets. BlueROV can also be used for many marine tasks, i.e., oyster monitoring\cite{lin2022oysternet} and navigation. Our group potentially wants to do marine-aerial navigation like Wang\cite{wang2018vision} has done for ground-aerial navigation. Detection of the BlueROV becomes necessary for such tasks. 

As we can see in Fig.~\ref{fig:ROV}, the 3D model in the simulation (Fig.~\ref{fig:ROV}(b)(d)) looks very similar to real BlueROV in the water (Fig.~\ref{fig:ROV}(a)(c)). Although there might be some visual difference between the images from the real data and the synthetic data, the actual object size in the UAV-based image is relatively small, so we can neglect the visual difference here. The relatively small size of the object in UAV-based images makes it possible for rendering the datasets in \textit{SeaDronesSim}, and directly use the datasets for training a network to detect the real object. 
\begin{figure}[t!]
		\centering      
		\includegraphics[width=\linewidth]{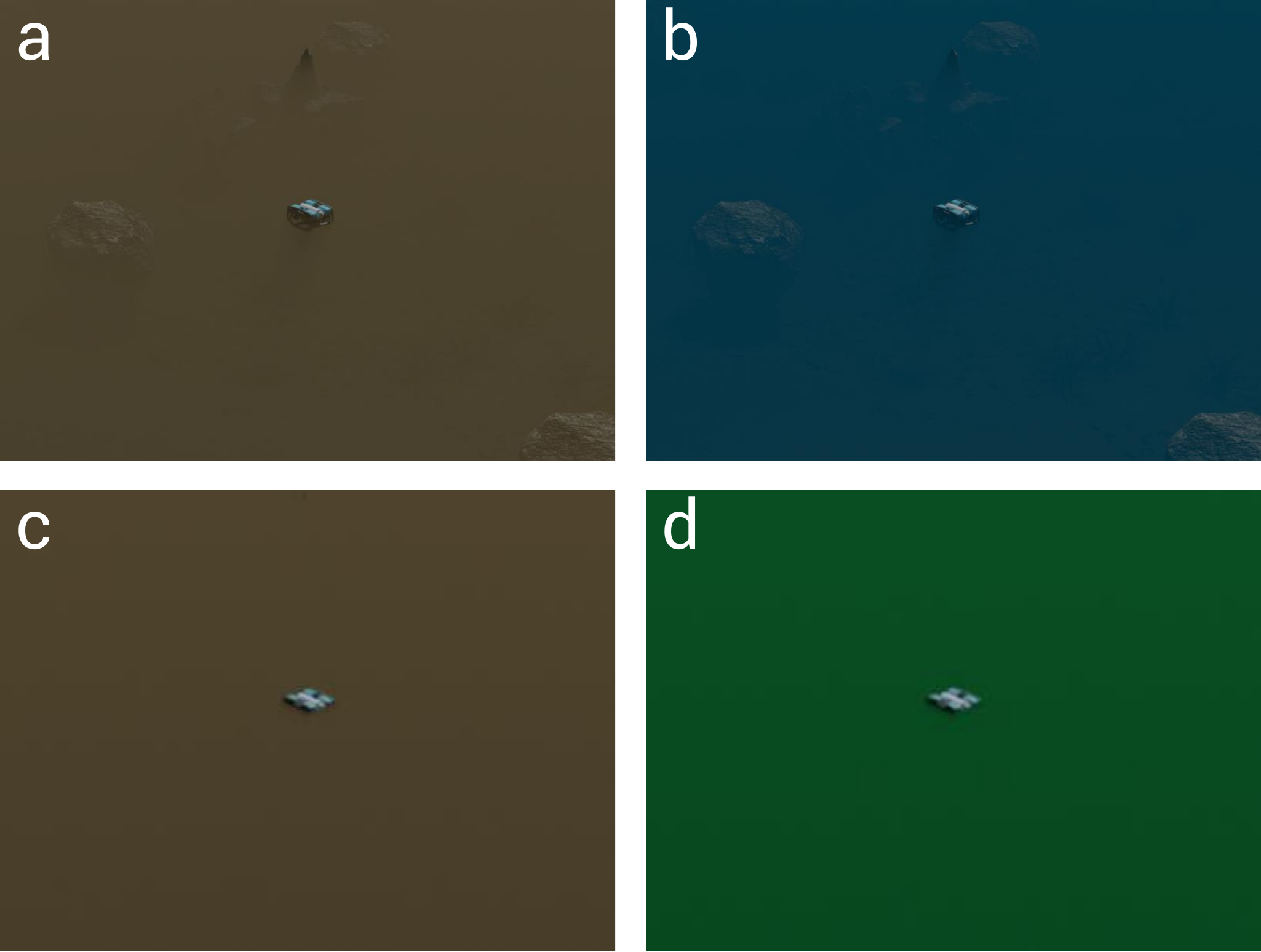}
		\caption{Different water colors and turbidity levels.}
		\label{fig:water_effect}
  \vspace{-4mm}
	\end{figure}
\subsection{Water Volume}
The major distinction between an on-land simulation and a maritime simulation is the water volume. In the Blender$^{\text{TM}}$ \cite{blender}environment, the render engine "CYCLE" uses path-tracing of the lights to generate images. In other words, it passes light with a certain number of bounces and receives light to generate images of different brightness, colors, and shades. By customizing particles in the water volume, we get a photo-realistic maritime image for different water colors and turbidity levels. The color of the water volume can also be modified to mimic a specific maritime environment's color scheme. 
As we can see in Fig.~\ref{fig:water_effect}(a) and Fig.~\ref{fig:water_effect}(b), these are UAV-based maritime images of different water colors in low turbidity. As we increase the turbidity level for the water, the object we added at the bottom (visible in Fig.~\ref{fig:water_effect}(a)) is no longer visible in Fig.~\ref{fig:water_effect}(c). Another image with different water colors in high turbidity is shown in Fig.~\ref{fig:water_effect}(d). 
\subsection{Aerial Image}
\begin{figure}[t!]
		\centering      
		\includegraphics[width=\linewidth]{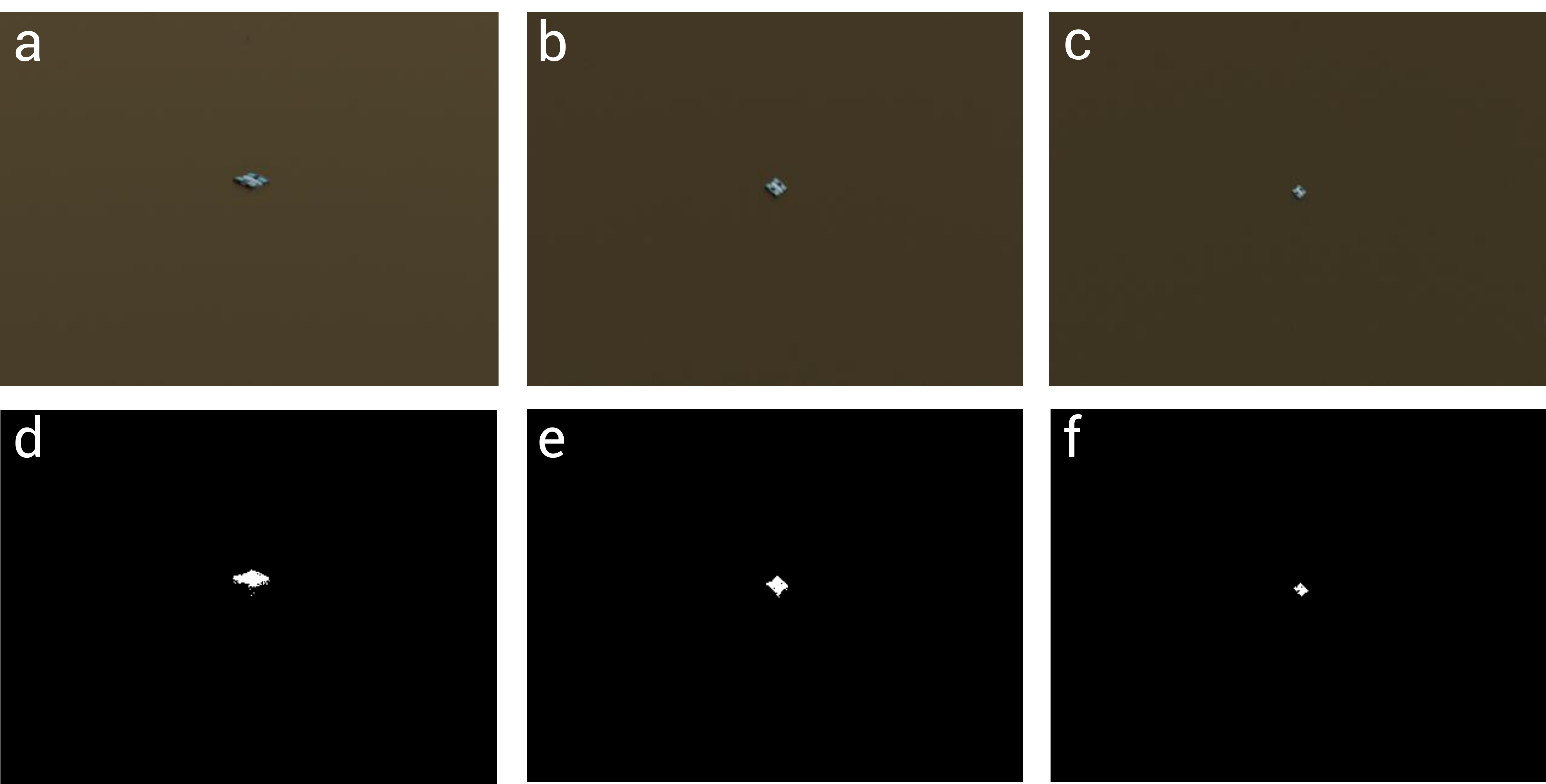}
		\caption{Images from different altitudes. (a) Image from an altitude equal to 20 m, (b) image from an altitude equal to 30 m, (c) image from an altitude equal to 40 m, (d)(e)(f) are the corresponding ground truth masks for (a)(b)(c).}
		\label{fig:train_depth}
  \vspace{-4mm}
	\end{figure}

We want to simulate the aerial Images in \textit{SeaDroneSim}. However, we do not care about the UAV's appearance and its physical model. The only thing that we care about is the camera attached to it. So we basically modeled the drone as a point object with a camera attached to it (we will just refer to the point object as the drone). Instead of moving the drone or the object in the simulation, we choose to move the object and fixate the camera on the object. In this case, when the object is closer to the drone, the object will look larger so that we can obtain multiple sizes within one run. However, this will not cover all possible sizes that we may want, so we adjust the altitude of the drone to obtain different images. As we can see in Fig.~\ref{fig:train_depth}, the object and its corresponding ground truth mask in the image become smaller when we increase the altitude when taking the images. The resolution of the image could go up to 30,000 x 30,000 pixels depending on the computational power of the hardware that is running the simulation. 
\subsection{Objects of Non-Interest}
To simulate some of the scenarios in that some objects not of interest are included, we add a few objects at the bottom of the seafloor and some hills above the sea level as we can see in Fig.~\ref{fig:sample_images}(a). 
\section{Experiments and Results}
\label{section:experiments_and_results}
We perform a comprehensive comparison with various training data by utilizing Faster R-CNN \cite{ren2015faster}. By rendering the training data in \textit{SeaDroneSim}, it can provide any number of images under different sizes, altitudes, angles, and colors. We resized the images to 416 x 416 pixels. For each masked image, the bounding boxes can be determined by the highest and the lowest x and y coordinates of BlueROV in the image. With these x and y coordinates, we added a BlueROV category and the corresponding bounding boxes. Therefore, each image receives a COCO-dataset format label. Instead of implementing a Faster R-CNN model from scratch, we used Detectron2 \cite{wu2019detectron2} which allows us to accelerate the developing process. Detectron2 is a novel system created by Facebook AI Research (FAIR) for computer vision research such as object detection and segmentation. 

\begin{figure}[t!]
		\centering      
		\includegraphics[width=\linewidth]{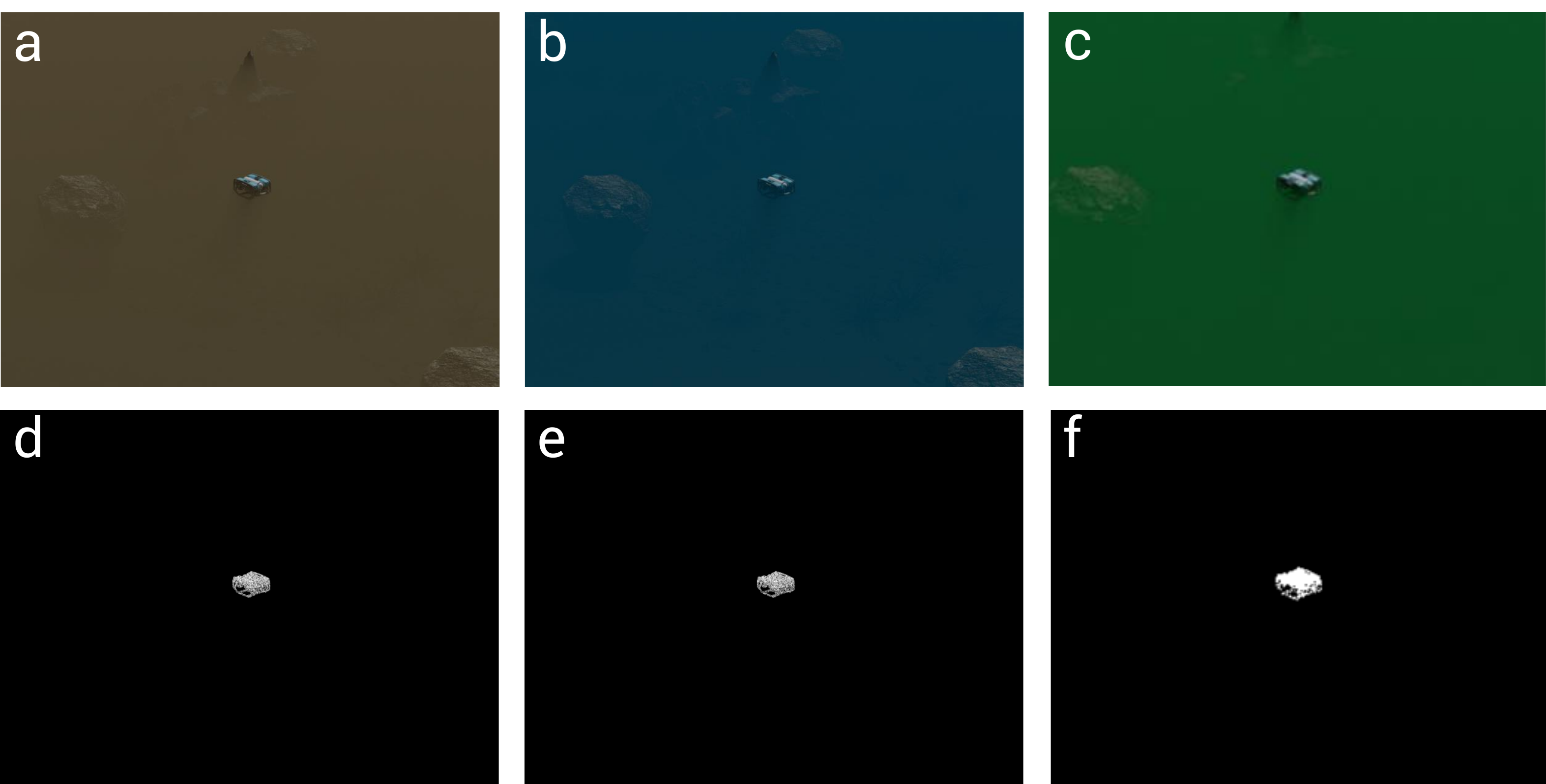}
		\caption{Examples of original and ground truth mask images. These examples are based on different background water colors brown (a, d), blue (b, e), and  green (c, f), respectively.}
		\label{fig:train_color}
  \vspace{-5mm}
	\end{figure}
\subsection{Synthetic Dataset}
There are 626 images for each dataset generated with the same parameters/constraints unless specified differently. For all the images from the simulation, we took 80\% for training and 20\% for validation. Fig.~\ref{fig:train_depth} shows the altitude images in the training dataset. To avoid the influence of noise, we removed the environment in the simulation. Therefore, it was not affected by other factors except for BlueROV. In the second part, we created a few datasets with different water colors in the simulation environment. There were 626 images for each color with multiple angles and altitudes in the maritime environment. Fig.~\ref{fig:train_color} shows the example images of different water colors in the training dataset. We could mimic the real UAV-based maritime images by modifying the water colors and adjusting the altitude for the image taken. 
\begin{figure*}[t!]
\begin{center}
    \includegraphics[width=1.5in]{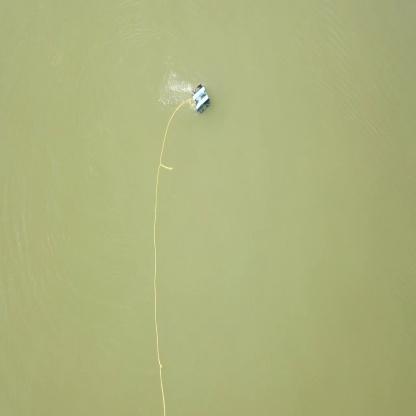}
    \includegraphics[width=1.5in]{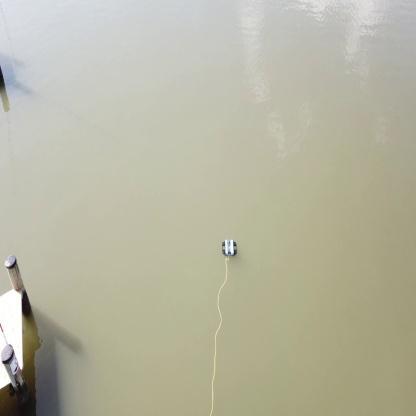}
    \includegraphics[width=1.5in]{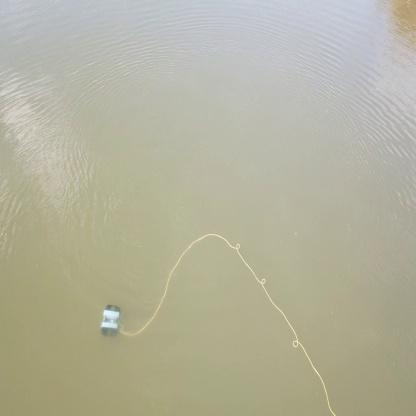}
    \includegraphics[width=1.5in]{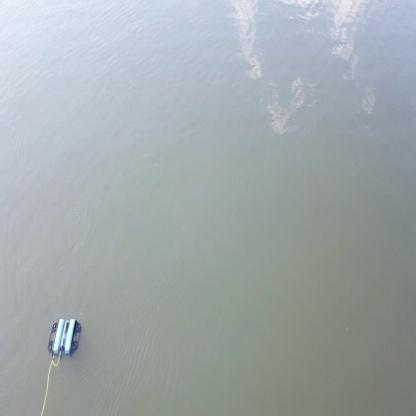}
    \includegraphics[width=1.5in]{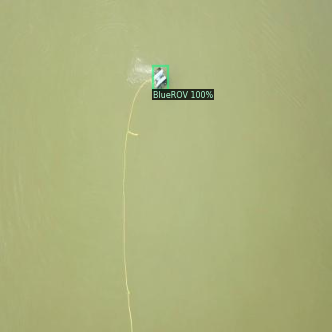}
    \includegraphics[width=1.5in]{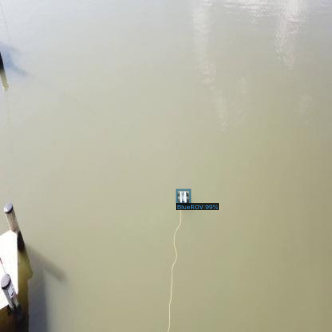}
    \includegraphics[width=1.5in]{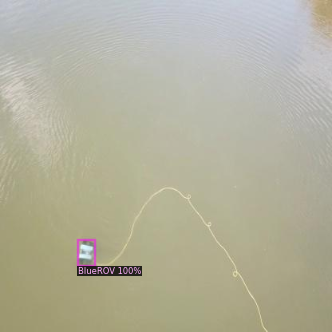}
    \includegraphics[width=1.5in]{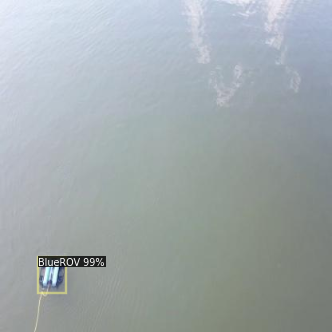}
\end{center}
    \caption{The first row is the examples of testing images. The second row is the testing results for the above images. These examples are collected from Sandy Point Park in Annapolis, MD.}
\label{fig:test}
\vspace{-5mm}
\end{figure*}
\subsection{Base model}
In this experiment, we used Faster R-CNN with X101-FPN as the backbone to train our dataset. Though X101-FPN takes more time and memory during training, it can produce higher accuracy results compared to other models in Faster R-CNN. In the model, we set up the learning rate of the model as 0.001. To prevent overfitting of the model to the synthetic data, the maximum number of iterations was 1,500, which can be adjusted according to the performance of validation. Since we only focused on detecting BlueROV in the image, there is only one class in this recognition task.

With dataset and model, it is convenient to train custom models in Detectron2. Registering the coco instance function, we uploaded training, validation, and testing data to the Detectron2 platform. Then we created a class and imported Faster R-CNN as our training model. A function exists that allows us to access the parameter of the model. From this function, we can assign the value to the parameter. The checkpoint and weight were saved in the output file. The checkpoint was provided to us to continue training the data after interrupting the training process. 

\subsection{Evaluation}
To prevent the convolutional neural network from overfitting to the synthetic dataset for a relatively easy task, we did not provide a large quantity of training data. However, these training data performed at least 95\% Average Precision \textit{(AP)} in the validation. Therefore, the next step is to evaluate the testing data. Testing data was collected from Sandy Point Park in Annapolis, MD. Fig.~\ref{fig:test} shows some of the real UAV-based maritime images. We used DJI pro2 [cite] and took a few video clips. We then carefully labeled 2,000 image frames selected from the video as the testing data to produce convincing results.
\begin{table}[b!]
\vspace{-3mm}
  \begin{center}
    \caption{The Average Precision \textit{(AP)} of the training data with various BlueROV image altitudes.}
    \label{tab:table1}
    \begin{tabular}{c|c|c|c|c|c} 
      \textbf{Altitude} & \textbf{$AP$} & \textbf{$AP_{50}$} & \textbf{$AP_{75}$} & \textbf{$AP_{s}$} & \textbf{$AP_{m}$}\\
      \hline
      10 m & 58.41 & 90.90 & 72.62 & 57.82 & 84.72\\
      20 m & 47.46 & 95.03 & 31.28 & 47.27 & 57.41\\
      30 m & 32.70 & 91.04 & 1.41 & 32.63 & 40.79\\
      40 m & 16.81 & 78.68 & 0.07 & 16.78 & 30.76\\
      50 m & 6.87 & 48.32 & 0.01 & 6.82 & 22.50\\
    \end{tabular}
  \end{center}
\end{table}

Table ~\ref{tab:table1} displays the output of the testing data with different training data obtained from various altitudes. Samples of images taken from altitudes 20 m, 30 m, and 40 m can be seen in Fig.~\ref{fig:train_depth}(a), Fig. ~\ref{fig:train_depth}(b), and Fig. ~\ref{fig:train_depth}(c), respectively. The highest mean Average Precision (mAP) is 58.41 at an altitude of 10 meters. Higher altitude results in a smaller object. Therefore, 10 meters altitude images contribute to the higher accuracy to match the testing data compared with others.

\begin{table}[b!]
\vspace{-3mm}
  \begin{center}
    \caption{The Average Precision \textit{(AP)} of the training data with different water colors at an image altitude of 10 m.}
    \label{tab:table2}
    \begin{tabular}{c|c|c|c|c|c} 
      \textbf{Water Color} & \textbf{$AP$} & \textbf{$AP_{50}$} & \textbf{$AP_{75}$} & \textbf{$AP_{s}$} & \textbf{$AP_{m}$}\\
      \hline
      Brown & 58.41 & 90.90 & 72.62 & 57.82 & 84.72\\
      Blue & 54.72 & 89.12 & 65.86 & 54.70 & 73.88\\
      Green & 57.71 & 89.24 & 71.33 & 57.51 & 78.73\\
    \end{tabular}
  \end{center}
  \vspace{-3mm}
\end{table}

Table ~\ref{tab:table2} presents the evaluation results for different water color training. As shown in Fig.~\ref{fig:train_color}, we applied three common water colors:  brown, blue, and green. Each color of training data contains different sizes and angles of the BlueROV, as mentioned in Sec.~\ref{section:proposed_approach}.3, to evaluate the result. Some of the results are displayed in the second row of Fig.~\ref{fig:test}. The brown training data achieved the highest mAP in Table ~\ref{tab:table2}, followed by blue, whereas green was much lower. This may be because the water color of our field data collected from Sandy Point Park was more brown, which may have made the training more robust to detect with a brown background. Alternatively, brown may simply provide a more distinguishable contrast color to that of the BlueROV, making is easier to detect. The reduced performance using the green dataset could be an issue with the model's ability to distinguish between the color of the BlueROV and the color of the maritime environment.


\begin{table}[b!]
  \vspace{-3mm}
  \begin{center}
    \caption{The Average Precision \textit{(AP)} of the training data with different data size at an image altitude of 10 m of Brown water color}
    \label{tab:table3}
    \begin{tabular}{c|c|c|c|c|c} 
      \textbf{Dataset Size} & \textbf{$AP$} & \textbf{$AP_{50}$} & \textbf{$AP_{75}$} & \textbf{$AP_{s}$} & \textbf{$AP_{m}$}\\
      \hline
      308 & 58.90 & 94.75 & 72.32 & 58.73 & 80.59\\
      626 & 58.41 & 90.90 & 72.62 & 57.82 & 84.72\\
      2500 & 45.38 & 95.52 & 25.13 & 45.06 & 76.88\\
      308+G & 71.00 & 96.42 & 87.24 & 70.94 & 80.08\\
      626+G & 64.87 & 98.27 & 77.75 & 65.04 & 70.69\\
    \end{tabular}
  \end{center}
\end{table}
We also perform a few experiments with different data sizes. We suspect that if the dataset is too large for our detection task, the network is likely overfitting to recognize objects in the synthetic domain. We use image capture for brown water color with an altitude equal to 10 meters. 
Then, we use the same network with the same number of iterations for training and only vary the images used in the training. As we can see, more data does not necessarily produce better results. A dataset of size 2500 is overfitting the network for the detection of the BlueROV. We also use 137 images generated from green water color, namely "G" in Table ~\ref{tab:table3}. Data from a different domain(water color) could improve the neural network's robustness and versatility. The result we obtained with a data size of 308 added G has 71 mAP which has a 20.5\% improvement over the result without adding different water color data. We have also performed training with only real data which is added in the appendix Table \ref{tab:table6}. Without actually training on any real data, we are able to get very close to the training with real dataset(77.56\% mAP). 
The same experiments for Table ~\ref{tab:table1} and Table ~\ref{tab:table2} are performed with more training data. These results are shown in the appendix. 

\vspace{-3mm}
\section{Conclusion and future work}
\vspace{-2mm}
\label{section:conclusion}
In this work, we utilized the capability of a game engine and built a simulation for rendering UAV-based maritime images. We discussed some of the implementation details for using Blender$^{\text{TM}}$ for generating synthetic datasets for object detection. One of the 3D models for the object of interest, BlueROV, is used as a feasibility study for the systematic automation approach for rendering UAV-based maritime images and then detection.  We then compared our detection result with the usage of different synthetic datasets generated from the simulation.  With only the usage of the generated synthetic dataset, we then set the benchmark for detecting BlueROV in maritime UAV-based images as 71.00 mAP. These results highlight that for data-critical applications when collecting real images is challenging, it is possible to use 3D model of the object to create photorealistic images that will successfully detect objects in a specific domain. This work is among the first research to build a UAV-based maritime image simulation focusing on object detection. In this new field, there are many directions and possible improvements that can be made to our \textit{SeaDroneSim} framework.  One possible future work would be obtaining the various 3D models for SeaDroneSee \cite{varga2022seadronessee} datasets and then rendering the datasets in \textit{SeaDroneSim}. It shall further increase  \textit{SeaDroneSim's} utilization scenarios and prove the practicality of our simulation. As for the detection phase, we adopted the existing network. New network architectures can be developed to tackle the UAV-based maritime domain problem. Finally, more object noise can be introduced into the simulation for more robust detection. Further, detection of objects from a single UAV could be applied to a swarm of UAVs units simultaneously, improving search capability and efficiency. Finally, detection angles from multiple UAVs could make the precise Geo-location of objects possible when traditional GPS methods are not possible. 
\section{ACKNOWLEDGMENT}
This work is supported by "Transforming Shellfish Farming with Smart Technology and Management Practices for Sustainable Production" grant no. 2020-68012-31805/project accession no. 1023149 from the USDA National Institute of Food and Agriculture.
{\small
\bibliographystyle{ieee_fullname}
\bibliography{egbib}
}

\appendix
\vspace{-3mm}
\section{\\Appendix A: More images and tables}
We perform more training with more images and we have put the result here in Appendix A. More data does not guarantee more accurate, it could cause the network to overfit to the synthetic dataset which would lower the accuracy. 
\begin{table}[h!]
  \begin{center}
    \caption{The Average Precision \textit{(AP)} of the training data(2500 images) with various BlueROV image altitudes}
    \label{tab:table4}
    \begin{tabular}{c|c|c|c|c|c} 
      \textbf{Altitude} & \textbf{$AP$} & \textbf{$AP_{50}$} & \textbf{$AP_{75}$} & \textbf{$AP_{s}$} & \textbf{$AP_{m}$}\\
      \hline
      10 m & 45.38 & 95.52 & 25.13 & 45.06 & 76.88\\
      20 m & 44.11 & 93.28 & 24.64 & 44.01 & 50.48\\
      30 m & 34.61 & 91.41 & 4.17  & 34.67 & 38.75\\
      40 m & 17.43 & 76.85 & 0.07  & 17.41 & 31.52\\
      50 m & 9.47 & 56.27 & 0.01  & 9.40 & 31.86\\
    \end{tabular}
  \end{center}
\end{table}

\begin{table}[h!]
\vspace{-8mm}
  \begin{center}
    \caption{The Average Precision \textit{(AP)} of the training data(2500 images) with different water colors at an image altitude of 10 m.}
    \label{tab:table5}
    \begin{tabular}{c|c|c|c|c|c} 
      \textbf{Water Color} & \textbf{$AP$} & \textbf{$AP_{50}$} & \textbf{$AP_{75}$} & \textbf{$AP_{s}$} & \textbf{$AP_{m}$}\\
      \hline
      Brown & 45.38 & 95.52 & 25.13 & 45.06 & 76.88\\
      Blue & 47.70 & 95.49 & 32.74 & 47.50 & 70.79\\
      Green & 45.67 & 95.93 & 25.68 & 45.39 & 74.11\\
    \end{tabular}
  \end{center}
  \vspace{-5mm}
\end{table}

\begin{table}[h!]
  \begin{center}
    \caption{Training with only real data with various training dataset sizes. We test with the remaining real dataset}
    \label{tab:table6}
    \begin{tabular}{c|c|c|c|c|c} 
      \textbf{Train Size} & \textbf{$AP$} & \textbf{$AP_{50}$} & \textbf{$AP_{75}$} & \textbf{$AP_{s}$} & \textbf{$AP_{m}$}\\
      \hline
      500 & 77.09 & 98.97 & 94.03 & 76.91 & 89.15\\
      1000 & 77.56 & 98.88 & 95.44 & 77.41 & 89.81\\
      1500 & 77.44 & 98.84 & 93.87 & 77.13 & 91.66\\
    \end{tabular}
  \end{center}
\end{table}
\end{document}